\documentclass[11pt]{article}

\usepackage[margin=1in]{geometry}
\usepackage{graphicx}
\usepackage{booktabs}
\usepackage{tabularx}
\usepackage{amsmath}
\usepackage[T1]{fontenc}
\usepackage[utf8]{inputenc}
\usepackage{microtype}
\usepackage{natbib}
\usepackage{xurl}
\usepackage{hyperref}

\hypersetup{
  colorlinks=true,
  citecolor=black,
  linkcolor=black,
  urlcolor=blue
}

\title{Measure Learning at Steady State\\
\large A {BIRD-SQL} Formula 1 case study}
\author{Manoj Bajaj}
\date{September 2026}

\begin{document}
\maketitle

\begin{abstract}
Continual Learning Bench (CL-Bench) scores learning as gain versus a stateless baseline on schedules of tens of instances, and reports that naive full-context in-context learning (ICL) is the strongest of the memory systems it tested. We keep ICL as \emph{an example} learning system, not as a claim that it is the best one, and we change the measure.

On a longer shared-world schedule, \textbf{steady-state learning} is the gap versus baseline on a pre-set late window, after exploration has plateaued. We also split the score into three curves that a single gain number mixes: \textbf{exploration efficiency} (SQL probes), \textbf{task reward} (binary hits), and \textbf{delivery cost} (API dollars and context size).

We study {BIRD-SQL} \texttt{formula\_1}: 174 analyst questions in fixed order on one Formula 1 database. Models are \texttt{gpt-5.6-luna} and \texttt{gpt-5.6-terra}. A fresh chat is the baseline; a resumed chat is ICL.

Learning is real, and it is largest late. On Luna, mean SQL probes on questions 135--174 fall from 4.6--5.6 (three baseline replicates) to 0.95 (ICL). That is about four to six probes saved per late question. Hits rise only modestly (74 vs.\ 65--67 of 174; late window 13 vs.\ 10--11 of 40). The same ICL transcript grows from $\sim$4k to $\sim$95k tokens, and API cost roughly doubles ($\sim$\$$9.92$ vs.\ $\sim$\$$4.47$--\$$4.55$). Terra repeats the pattern: late probes 1.08 vs.\ 3.85, hits 87 vs.\ 63, cost \$$11.10$ vs.\ \$$3.63$, context $\sim$3k to $\sim$103k.

Short-horizon gain, of the kind CL-Bench can report, understates the late exploration saving and never sees the cost curve invert. Unbounded ICL is therefore a poor candidate for ``the'' learning mechanism: it helps, then it taxes every later job.
\end{abstract}

\section{Introduction}

A production agent that answers many questions against one database should get cheaper to operate as it learns the schema. CL-Bench~\citep{asawa2026clbench} made this comparison concrete: a stateful system versus its own reset baseline, on expert-validated environments. Its headline is that naive ICL beat Mem0, ACE, and ICL Notepad on most of those short schedules. Peak normalized gain was 25.4\%. Sequences were on the order of tens of instances. The Database Exploration task is 40 questions with a schema change at question 20.

Two problems follow if that protocol is treated as the definition of agent learning.

First, \textbf{the interesting saving often appears after the short schedule ends.} Early jobs still explore. A mean over the first 20 or 40 questions mixes that transient with later reuse. We instead pre-set a late window (last 40 of 174) and call the baseline gap there \textbf{steady-state learning}.

Second, \textbf{one gain number cannot tell whether the system got more accurate, more efficient, or more expensive.} On this task those three stories disagree. We report them separately:
\begin{enumerate}
  \item \textbf{Exploration efficiency} --- SQL probes (\texttt{db query}) per question.
  \item \textbf{Task reward} --- binary correctness.
  \item \textbf{Delivery cost} --- API dollars and prompt size.
\end{enumerate}

We use ICL because it is the simple system CL-Bench found strongest, and because it is easy to ablate (keep the chat, or start a new one). We do \textbf{not} claim ICL is the best learning mechanism. Better memory, retrieval, or trained compactors should exist. The case study is here to show that learning systems help, that the right score is the late gap, and that ranking ICL as best is unsafe once context is allowed to grow without bound.

Public traces are on Harbor Hub~\citep{harborhub2026}. Interactive per-question plots for Luna r1 are an ancillary HTML file (not the article itself).

\section{Related work}

\paragraph{CL-Bench.}
\citet{asawa2026clbench} introduce Continual Learning Bench: six stateful environments and a gain metric versus each system's own stateless baseline. Naive full-context ICL outperforms the dedicated memory architectures they evaluated on most tasks. Accumulated state often hurts those memories (spurious generalizations, stale beliefs). Higher spend does not reliably buy higher gain. They already note that sequences of tens of instances are shorter than many deployment horizons. We take the reset comparison and the shared-world design. We change the horizon and the aggregation: late-window curves instead of a short-run mean, and three measures instead of one.

We do not re-run Mem0 or ACE. We do not claim those systems would lose on Formula 1. We do claim that calling ICL the best \emph{mechanism} does not survive a long transcript: the same state channel that stores schema knowledge also grows input tokens and dollars.

\paragraph{ARC-AGI-3.}
\citet{arcprize2026arcagi3} treat intelligence as skill-acquisition efficiency. We apply that efficiency lens to repeated work in one database (probes and dollars), not to novel game episodes.

\paragraph{AgentCL.}
\citet{shu2026agentcl} argue for controls, ordering, and isolating learning from base skill. That is why we keep a per-question reset baseline, fix question order, and report the late window separately from the whole-run mean.

\paragraph{Prompt caching.}
Provider caches make early resumed turns look cheap: the shared prefix is billed as a cache read. Caches do not remove the need to send a longer prefix later. Cost curves must be read with context size, not only with cache-hit rate. Related compact representations include Cartridges~\citep{eyuboglu2025cartridges}. Native conversation summarization is an experiment in Section~\ref{sec:compact}, not a proposed system.

\paragraph{BIRD-SQL.}
The environment is the Formula 1 database and question bank from BIRD~\citep{li2023bird}. We use the SQLite file from BIRD Mini-Dev and the \texttt{formula\_1} subset of the 2025-11-06 development dump (174 questions in official order). This is a long-horizon case study, not a new text-to-SQL leaderboard.

\section{Setup}

\subsection{Task}

BIRD-SQL \texttt{formula\_1}: one SQLite database, \textbf{174} natural-language questions in the official fixed order. The agent may run SQL only through \texttt{db query} and must write an answer file. Correctness is binary. The schema does not change mid-run (unlike CL-Bench Database Exploration).

\subsection{Conditions}

\begin{center}
\begin{tabular}{ll}
\toprule
Condition & Session \\
\midrule
Baseline & Fresh chat each question \\
ICL & Same chat resumed (\texttt{--resume-trajectory}) \\
\bottomrule
\end{tabular}
\end{center}

Harbor \texttt{pi}, \texttt{--n-concurrent 1}. Independent jobs are separate rollouts; Harbor has no seed flag.

A third arm, native pi compaction at $\geq$25k tokens, is reported only as a negative experiment (Section~\ref{sec:compact}). It is not part of the main claim.

\subsection{Models and jobs}

\begin{center}
\small
\begin{tabular}{lll}
\toprule
Model & Baseline & ICL \\
\midrule
\texttt{openai/gpt-5.6-luna} & 3 full runs (r1--r3) & 1 full run \\
\texttt{openai/gpt-5.6-terra} & 1 full run & 1 full run (q1--121, then seeded resume q122--174) \\
\bottomrule
\end{tabular}
\end{center}

Terra ICL is one trajectory with a seam at question 122: the verifier used to abort on mixed-shape answers (\texttt{[["STR","VER","OCO"], 1]}). After the fix, the q-121 session was continued. Hits and probes are stitched; dollars are the sum of the two Harbor jobs.

\subsection{Metrics}

\begin{center}
\small
\begin{tabularx}{\linewidth}{lX}
\toprule
Name & Definition \\
\midrule
Hits & Binary correctness \\
Queries & \texttt{db query} calls per question \\
Steady-state learning & Baseline minus learner, on questions \textbf{135--174} (pre-set last 40). Primary: mean queries. Hits and dollars on the same window. \\
Context tokens & Estimated prompt size (input + cache-read) \\
API cost & Provider USD \\
\bottomrule
\end{tabularx}
\end{center}

Whole-run means mix early exploration with the plateau. They are secondary. We do not report wall time: it is a property of our runner, retries, and machine, not of the learning system.

\section{Learning systems help, especially at the end}

\subsection{Whole run versus late window}

\begin{table}[t]
\centering
\caption{Whole run ($n=174$).}
\label{tab:whole}
\small
\begin{tabular}{lrrrr}
\toprule
Run & Hits & Hit rate & Mean queries & Cost (USD) \\
\midrule
Luna baseline r1 & 65 & 37.4\% & 4.78 & 4.55 \\
Luna baseline r2 & 67 & 38.5\% & 4.40 & 4.51 \\
Luna baseline r3 & 65 & 37.4\% & 4.43 & 4.47 \\
Luna ICL & 74 & 42.5\% & 1.07 & 9.92 \\
Terra baseline & 63 & 36.2\% & 3.76 & 3.63 \\
Terra ICL & 87 & 50.0\% & 1.17 & 11.10 \\
\bottomrule
\end{tabular}
\end{table}

ICL uses about \textbf{one SQL probe per question} after the first few jobs (Table~\ref{tab:whole}). Baseline keeps issuing about four. Hits move in the same direction, but the hit gap is small on Luna ($+$7 to $+$9 of 174) and larger on Terra ($+$24). If the only score were accuracy, Luna would look like a weak learning result. The probe gap is not weak.

\begin{table}[t]
\centering
\caption{Steady-state window (questions 135--174).}
\label{tab:late}
\small
\begin{tabular}{lrrr}
\toprule
Run & Hits / 40 & Mean queries & Window cost (USD) \\
\midrule
Luna baseline r1 & 11 & \textbf{5.58} & 1.16 \\
Luna baseline r2 & 11 & \textbf{4.60} & 1.09 \\
Luna baseline r3 & 10 & \textbf{5.25} & 1.18 \\
Luna ICL & 13 & \textbf{0.95} & 3.56 \\
Terra baseline & 9 & \textbf{3.85} & 0.91 \\
Terra ICL & 19 & \textbf{1.08} & --- \\
\bottomrule
\end{tabular}
\end{table}

\paragraph{Steady-state exploration learning (Luna).}
Late probes are 0.95 vs.\ 4.6--5.6 (Table~\ref{tab:late}). Relative to the mean of the three baselines (5.14), ICL saves about 4.2 SQL probes per late question, or on the order of 170 probes over the last 40 questions. Terra saves about 2.8 probes per late question (1.08 vs.\ 3.85).

Late hits on Luna barely move (13 vs.\ 10--11). Terra ICL does pick up late hits (19 vs.\ 9). Exploration efficiency is the reliable late signal across both models; reward is model-dependent.

Cumulative probes on Luna r1: 186 (ICL) vs.\ 832 (baseline).

\subsection{The curve, not the average}

\begin{figure}[t]
\centering
\includegraphics[width=\linewidth]{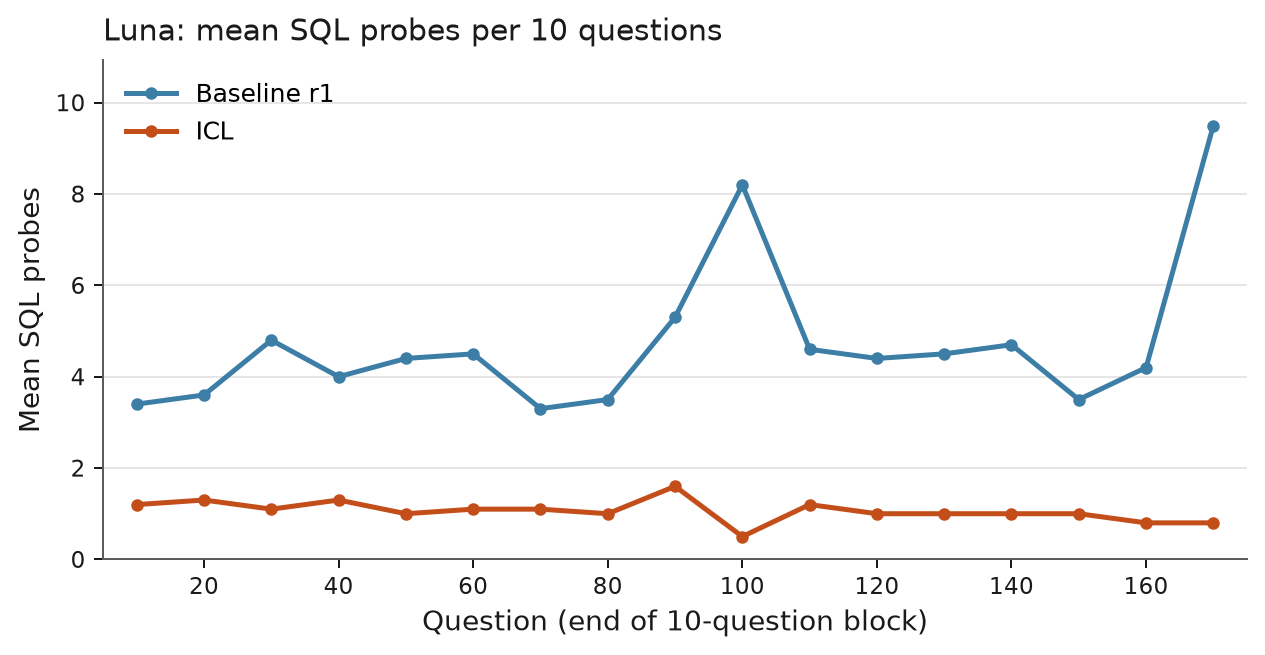}
\caption{Luna r1. Each point is the mean of ten consecutive questions (last plotted block is q161--170). ICL stays near one probe. Baseline does not.}
\label{fig:luna-probes}
\end{figure}

Figure~\ref{fig:luna-probes} is mean SQL probes in successive blocks of ten questions (Luna r1). ICL is near 1 from the second block onward. Baseline does not fall. The late gap is not a slow climb that a 40-question study would catch in miniature. It is a plateau versus a system that never stops exploring.

\begin{figure}[t]
\centering
\includegraphics[width=\linewidth]{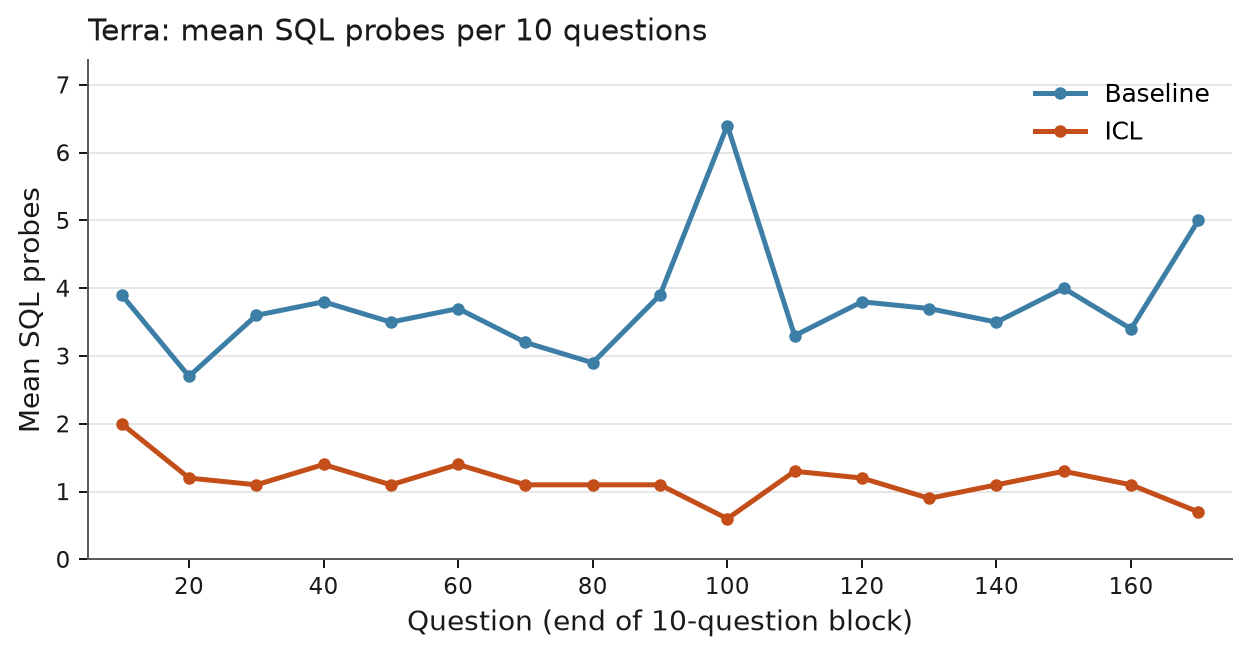}
\caption{Terra. Same block means. ICL again sits near one probe after the first block.}
\label{fig:terra-probes}
\end{figure}

\begin{figure}[t]
\centering
\includegraphics[width=\linewidth]{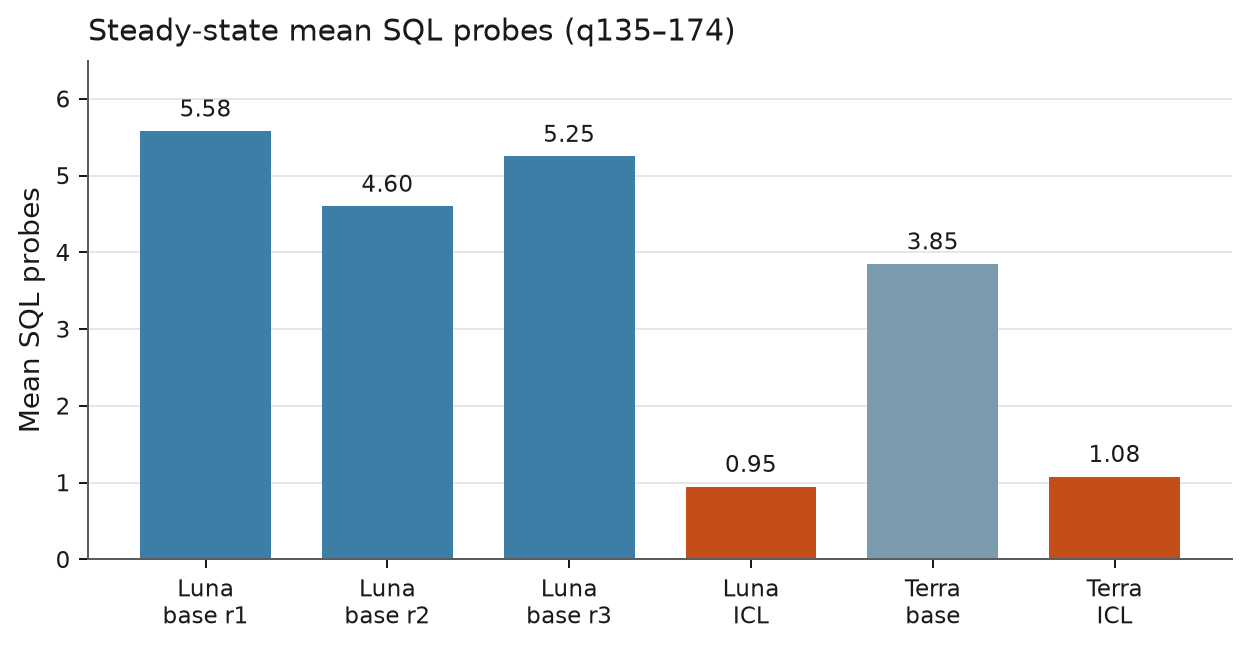}
\caption{The measure we want: late-window mean probes. ICL is near 1 on both models. Baseline stays in the 4--6 (Luna) or $\sim$4 (Terra) range.}
\label{fig:steady}
\end{figure}

Time-to-plateau is separate from steady-state learning. On Luna, ICL is already near one probe by question 2. The late window is not ``when learning starts.'' It is the period where the saving is still there, after a short eval would have stopped.

\section{Why this measure, and why not ``ICL is best''}

\subsection{Short-horizon gain hides the late saving}

CL-Bench Database Exploration is 40 questions. Our first 40 Luna questions are the closest analogue (no schema migration).

\begin{table}[t]
\centering
\caption{Luna r1, first 40 vs.\ last 40.}
\label{tab:windows}
\small
\begin{tabular}{lrrrrr}
\toprule
Window & Base hits & ICL hits & Base $q$ & ICL $q$ & Probe cut \\
\midrule
q1--40 (CL-Bench length) & 19 & 22 & 3.95 & 1.23 & 69\% \\
q135--174 (steady state) & 11 & 13 & 5.58 & 0.95 & 83\% \\
\bottomrule
\end{tabular}
\end{table}

A 40-question mean already shows ICL helping (Table~\ref{tab:windows}). It \textbf{understates} the late exploration gap (baseline probes are \emph{higher} late, ICL is \emph{lower}). It also reports a hit story that barely moves. If the product goal is ``stop rediscovering the schema,'' the last 40 questions are the score. Averaging the first 40 with the rest, or stopping at 40, dilutes it.

Whole-run mean queries (1.07 vs.\ 4.78) sit between the two windows. That mixed number is what a single-run ``gain'' on 174 questions would still smear.

\subsection{Three measures, three conclusions}

On Luna r1:

\begin{center}
\small
\begin{tabularx}{\linewidth}{lX}
\toprule
If you score\ldots & You conclude\ldots \\
\midrule
Hits & Small whole-run lift (74 vs.\ 65) \\
SQL probes & Large whole-run cut (1.07 vs.\ 4.78); \textbf{larger} late (0.95 vs.\ 5.58) \\
API dollars & ICL \textbf{loses}: \$$9.92$ vs.\ \$$4.55$ whole run; \$$3.56$ vs.\ \$$1.16$ late \\
\bottomrule
\end{tabularx}
\end{center}

CL-Bench's single gain, and a short schedule, cannot produce this split. That is the measurement change.

\subsection{Unbounded context makes ICL a costly mechanism}

\begin{figure}[t]
\centering
\includegraphics[width=\linewidth]{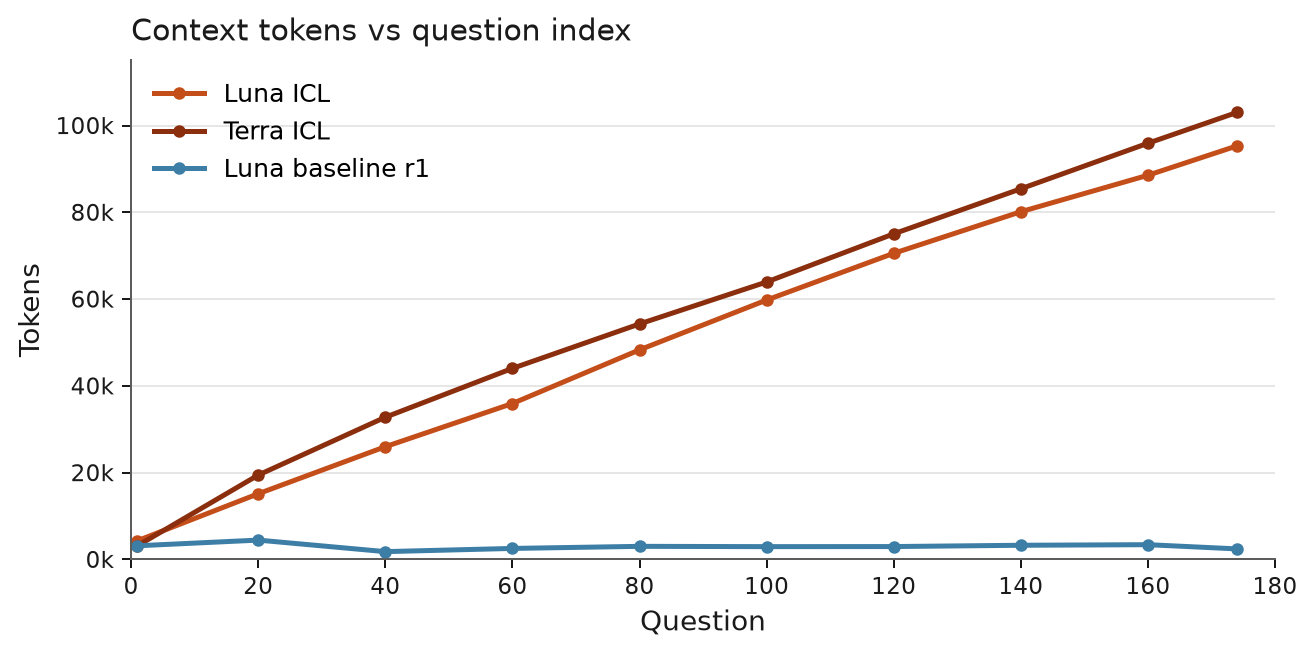}
\caption{ICL context (input + cache-read). Luna ends at $\sim$95k tokens, Terra at $\sim$103k. Baseline is a new prompt each time.}
\label{fig:context}
\end{figure}

Figure~\ref{fig:context} is prompt size. ICL is linear in the transcript. Baseline stays a few thousand tokens per question (a new chat).

\begin{figure}[t]
\centering
\includegraphics[width=\linewidth]{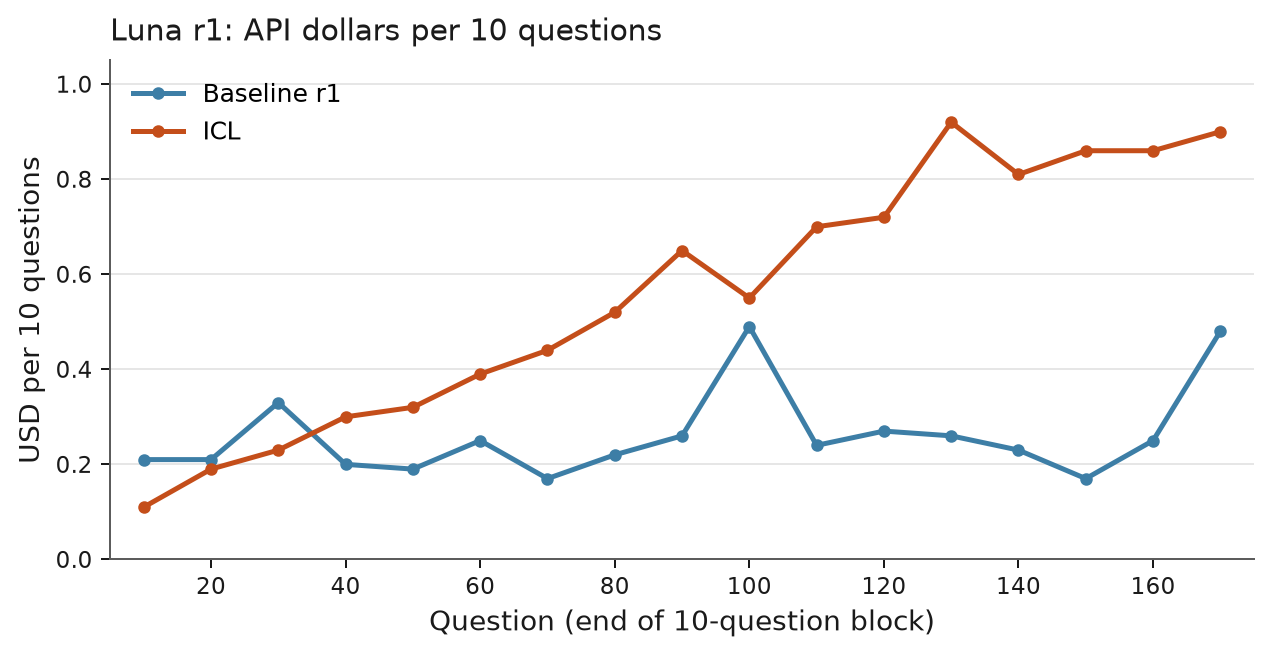}
\caption{Delivery cost. ICL starts cheaper, then the curve rises with context. Baseline stays roughly flat. This inversion is invisible on a 10- or 40-question schedule.}
\label{fig:cost}
\end{figure}

Figure~\ref{fig:cost} is API dollars per ten questions (Luna r1). Early ICL is cheaper than baseline (cache on a short prefix). By mid-run ICL is more expensive, while SQL probes stay at one. Late blocks cost about \$$0.85$--\$$0.92$ per ten ICL questions versus about \$$0.17$--\$$0.48$ for baseline. Prompt cache is on; the cached prefix is just longer.

This is the refutation of ``ICL is the best learning mechanism,'' not of ``ICL beat the memories CL-Bench tested on short tasks.'' Full context is a working store for schema knowledge. It is also an input bill that grows with every later job. A better learning system would keep the late probe gap \textbf{without} shipping 95k tokens on question 174.

Terra ICL total spend (\$$11.10$) versus Terra baseline (\$$3.63$) is the same qualitative result.

\subsection{Experiment: native compaction does not fix delivery}
\label{sec:compact}

We forced pi's default \texttt{ctx.compact()} at $\geq$25k tokens on the same resumed session (no custom summarizer). Luna: 134 successful compactions, one summarizer failure, context held near 25--32k (q1 $\sim$4.1k, q174 $\sim$26.5k). Terra: 141 compactions, context $\sim$29k at the end.

\begin{table}[t]
\centering
\caption{Compact versus ICL (full 174).}
\label{tab:compact}
\small
\begin{tabular}{lrrrrr}
\toprule
Run & Hits & Mean $q$ & Late $q$ & Cost (USD) & End ctx \\
\midrule
Luna ICL & 74 & 1.07 & 0.95 & 9.92 & $\sim$95k \\
Luna compact 25k & 71 & 1.14 & 1.18 & \textbf{12.73} & $\sim$26k \\
Terra ICL & 87 & 1.17 & 1.08 & 11.10 & $\sim$103k \\
Terra compact 25k & 81 & 1.16 & 1.25 & \textbf{14.21} & $\sim$29k \\
\bottomrule
\end{tabular}
\end{table}

Native compaction keeps probes near ICL and bounds tokens (Table~\ref{tab:compact}). It does \textbf{not} cut dollars; it costs more. Untuned summarization, broken cache prefixes after each compact, and extra summarizer calls are enough to explain that. Fine-tuned or learned compactors might do better. This paper does not claim to have found one. Native compact is not a substitute for a real learning system.

\section{Limitations}

One database, one official question order, no mid-run schema change. One ICL rollout per model; Luna baseline has three replicates, ICL variance is unmeasured. Terra ICL has a resume seam at q122. ICL is the only learning mechanism in the main tables. Dollars follow provider usage in Harbor. Two OpenAI \texttt{gpt-5.6-*} models.

\section{Conclusion}

On Formula 1, a simple learning system helps, and the help is largest where it should be scored: \textbf{at steady state}. Late SQL probes fall to about one per question versus four to six for a fresh chat. That is schema reuse, not an accuracy miracle.

CL-Bench's short-horizon gain is the wrong aggregator for this fact. A 40-question mean already shows a probe cut and almost no hit lift. It shrinks the late exploration gap and never watches cost. Once the transcript is allowed to grow, ICL's API bill rises while probes stay flat. That is why ICL should not be treated as the best learning mechanism, even if it won CL-Bench's short memory comparison.

The measurement we want is boring to state: late-window gap versus baseline, on exploration, reward, and delivery cost, separately. A system that keeps the first curve without the third is the actual target.

\appendix
\section{Jobs and traces}
\label{app:jobs}

Traces are public on Harbor Hub~\citep{harborhub2026}.
The ancillary file \texttt{harbor-hub-urls.txt} lists every job.
Pages use the path \texttt{/jobs/}\textit{uuid} on that site.

{\small
\raggedright
\begin{tabularx}{\linewidth}{l>{\ttfamily\raggedright\arraybackslash}X}
\toprule
Arm & Job UUID \\
\midrule
Luna baseline r1 & \href{https://hub.harborframework.com/jobs/38429c53-e925-4915-be86-be3d6b5f7a91}{38429c53-e925-4915-be86-be3d6b5f7a91} \\
Luna baseline r2 & \href{https://hub.harborframework.com/jobs/3bc81708-5219-40fe-b485-0128b2fb5a8a}{3bc81708-5219-40fe-b485-0128b2fb5a8a} \\
Luna baseline r3 & \href{https://hub.harborframework.com/jobs/3824a4cc-7e8e-4d41-bd09-452ed6187566}{3824a4cc-7e8e-4d41-bd09-452ed6187566} \\
Luna ICL & \href{https://hub.harborframework.com/jobs/c5a462ac-ed07-4d77-895b-00544701970a}{c5a462ac-ed07-4d77-895b-00544701970a} \\
Luna compact 25k & \href{https://hub.harborframework.com/jobs/06552de1-23d7-411e-bbf5-c093ff35b639}{06552de1-23d7-411e-bbf5-c093ff35b639} \\
Terra baseline & \href{https://hub.harborframework.com/jobs/fe2a8bdb-ef3d-40a9-b696-ff43267f4801}{fe2a8bdb-ef3d-40a9-b696-ff43267f4801} \\
Terra ICL q1--121 & \href{https://hub.harborframework.com/jobs/38e0271b-0ca4-4e7d-be85-e0db8bc3ed5a}{38e0271b-0ca4-4e7d-be85-e0db8bc3ed5a} \\
Terra ICL q122--174 & \href{https://hub.harborframework.com/jobs/eedac1a3-6d8d-4389-b66b-606b98b75ddd}{eedac1a3-6d8d-4389-b66b-606b98b75ddd} \\
Terra compact 25k & \href{https://hub.harborframework.com/jobs/3153e228-23b5-4beb-9398-d7d341797302}{3153e228-23b5-4beb-9398-d7d341797302} \\
\bottomrule
\end{tabularx}
\par}

The Luna r1 interactive viewer is \texttt{anc/f1-luna-r1-icl-vs-baseline.html}. Stitched Terra ICL metrics are \texttt{anc/f1-icl-full-r2-terra-stitched.json}.

\bibliographystyle{plainnat}
\bibliography{refs}

@article{li2023bird,
  title={{Can {LLM} Already Serve as A Database Interface? A {BIg} Bench for Large-Scale Database Grounded Text-to-{SQL}s}},
  author={Li, Jinyang and Hui, Binyuan and Qu, Ge and Yang, Jiaxi and Li, Binhua and Li, Bowen and Wang, Bailin and Qin, Bowen and Cao, Rongyu and Geng, Ruiying and Huo, Nan and Zhou, Xuanhe and Ma, Chenhao and Li, Guoliang and Chang, Kevin C. C. and Huang, Fei and Cheng, Reynold and Li, Yongbin},
  journal={Advances in Neural Information Processing Systems},
  volume={36},
  year={2023},
  eprint={2305.03111},
  archivePrefix={arXiv},
  primaryClass={cs.DB},
  url={https://arxiv.org/abs/2305.03111}
}

@misc{asawa2026clbench,
  title={{Continual Learning Bench: Evaluating Frontier {AI} Systems in Real-World Stateful Environments}},
  author={Asawa, Parth and Glaze, Christopher M. and Orlanski, Gabriel and Ramakrishnan, Ramya and Xu, Benji and Biswal, Asim and Chen, Vincent Sunn and Sala, Frederic and Zaharia, Matei and Gonzalez, Joseph E.},
  year={2026},
  eprint={2606.05661},
  archivePrefix={arXiv},
  primaryClass={cs.AI},
  url={https://arxiv.org/abs/2606.05661}
}

@misc{arcprize2026arcagi3,
  title={{ARC-AGI-3}: A New Challenge for Frontier Agentic Intelligence},
  author={{ARC Prize Foundation}},
  year={2026},
  eprint={2603.24621},
  archivePrefix={arXiv},
  primaryClass={cs.AI},
  url={https://arxiv.org/abs/2603.24621}
}

@misc{shu2026agentcl,
  title={{AgentCL}: Toward Rigorous Evaluation of Continual Learning in Language Agents},
  author={Shu, Yiheng and Guti{\'e}rrez, Bernal Jim{\'e}nez and Jonnalagedda, Saisri Padmaja and Yao, Yuguang and Sun, Huan and Su, Yu},
  year={2026},
  eprint={2606.02461},
  archivePrefix={arXiv},
  primaryClass={cs.AI},
  url={https://arxiv.org/abs/2606.02461}
}

@misc{eyuboglu2025cartridges,
  title={{Cartridges: Lightweight and general-purpose long context representations via self-study}},
  author={Eyuboglu, Sabri and Ehrlich, Ryan and Arora, Simran and Guha, Neel and Zinsley, Dylan and Liu, Emily and Tennien, Will and Rudra, Atri and Zou, James and Mirhoseini, Azalia and R{\'e}, Christopher},
  year={2025},
  eprint={2506.06266},
  archivePrefix={arXiv},
  primaryClass={cs.CL},
  url={https://arxiv.org/abs/2506.06266}
}

@misc{harborhub2026,
  title={{Harbor Hub}},
  author={{Harbor Hub}},
  howpublished={\url{https://hub.harborframework.com/}},
  year={2026},
  note={Public traces: \url{https://hub.harborframework.com/jobs/3153e228-23b5-4beb-9398-d7d341797302}}
}

\end{document}